\documentclass[11pt]{article}

\usepackage[preprint]{acl}

\usepackage{times}
\usepackage{latexsym}
\usepackage[T1]{fontenc}
\usepackage[utf8]{inputenc}
\usepackage{microtype}
\usepackage{inconsolata}

\usepackage{graphicx}
\usepackage{subcaption}
\usepackage{booktabs}
\usepackage{multirow}
\usepackage{placeins}

\usepackage{amsmath}
\usepackage{amsfonts}

\usepackage{url}
\usepackage{xcolor}

\definecolor{todored}{rgb}{0.6,0,0}
\newif\ifshowtodos
\showtodostrue

\title{RefinedGAW: A Refined German Academic Web Dataset}

\author{
  \textbf{Maximilian Idahl}\textsuperscript{1,2} \quad
  \textbf{Oleh Astappiev}\textsuperscript{2} \quad
  \textbf{Benedikt Droste}\textsuperscript{1} \quad
  \textbf{Robert Scholz}\textsuperscript{1} \\
  \textbf{Björn Plüster}\textsuperscript{1} \quad
  \textbf{Jan Philipp Harries}\textsuperscript{1} \quad
  \textbf{Robert Jäschke}\textsuperscript{3} \quad
  \textbf{Wolfgang Nejdl}\textsuperscript{2} \\[3pt]
  \normalfont\textsuperscript{1}ellamind \\
  \normalfont\textsuperscript{2}L3S Research Center, Leibniz University Hannover\\
  \normalfont\textsuperscript{3}Berlin School of Library and Information Science, Humboldt-Universität zu Berlin
}

\begin{document}
\maketitle

\begin{abstract}
The largest open German text datasets are built from broad web crawls, which provide scale but limited coverage of specific domains.
We introduce RefinedGAW, a processed and annotated resource derived from 26 semi-annual crawls of the German Academic Web, totaling 122 million documents and 314 billion tokens, with full crawl and capture history of every text.
Documents labeled as academic research account for 84\% of German RefinedGAW tokens across HTML and PDF.
At 87.6 billion tokens, this exceeds the 64.9 billion tokens in the German subsets of FineWeb2 and FinePDFs, despite their almost five-times-larger combined size.
RefinedGAW also contains substantial amounts of educational, math-heavy, and expert-oriented content with minimal commercial bias.
Across HTML and PDF, 88\% of German RefinedGAW documents have no near-duplicate candidate match in the three evaluated reference datasets, supporting complementary coverage.
We release a queryable metadata index for fine-grained data selection and a rehydration tool for retrieving and processing source documents.
\end{abstract}

\section{Introduction}
\label{sec:introduction}
Large language models have increased demand for text datasets at substantial scale.
The largest open text datasets are derived from broad web crawls, such as Common Crawl~\citep{commoncrawl}.
Focused crawls can complement these resources by covering specific domains more exhaustively, surfacing specialized material.

\begin{figure*}[t]
    \centering
    \begin{subfigure}[t]{0.48\textwidth}
        \centering
        \includegraphics[width=\linewidth,height=1.7in,keepaspectratio]{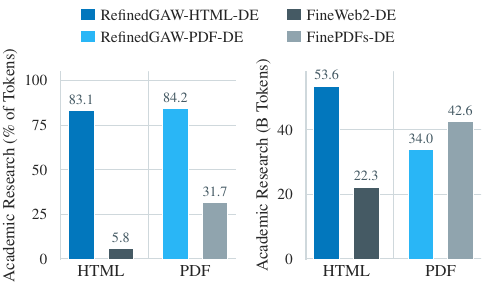}
        \caption{Academic-research concentration and yield.}
        \label{fig:refinedgaw-academic}
    \end{subfigure}
    \hfill
    \begin{subfigure}[t]{0.48\textwidth}
        \centering
        \includegraphics[width=\linewidth,height=1.7in,keepaspectratio]{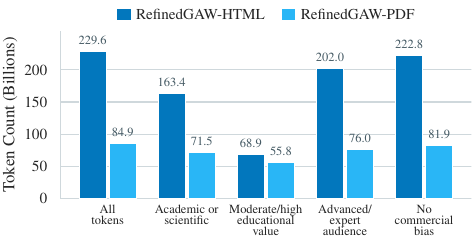}
        \caption{Selected token yields in RefinedGAW.}
        \label{fig:refinedgaw-yields}
    \end{subfigure}
    \caption{Content composition and token yields in RefinedGAW.
    Panel (a) reports the share and count of tokens in documents labeled as academic research across German HTML and PDF datasets.
    Panel (b) reports token yields for selected content labels in the complete RefinedGAW HTML and PDF collections.
    }
    \label{fig:refinedgaw-headline}
\end{figure*}

The German Academic Web\footnote{\url{https://german-academic-web.de/}}~(GAW) archive is a domain-specific crawling effort containing semi-annual snapshots of the public web presence of German academic institutions from 2012 onwards~\citep{paris2020gaw}, starting from around 150 seed domains.
A potentially high concentration of scientific and educational text, which improves language-model performance on knowledge- and reasoning-intensive benchmarks~\citep{penedo2024fineweb}, makes academic websites a promising domain.
However, raw web archive captures require substantial processing before they can support downstream research and model development.
Resources such as C4, mC4, RefinedWeb, MADLAD-400, FineWeb, and HPLT illustrate the extensive processing and documentation needed to turn web captures into usable text datasets~\citep{raffel2020t5,xue2021mt5,penedo2023refinedweb,kudugunta2023madlad,penedo2024fineweb,oepen2026hplt}.

With RefinedGAW\footnote{Available at \url{https://hf.co/datasets/ellamind/RefinedGAW}.}, we introduce a processed and richly annotated resource derived from the GAW.
In two separate pipelines we extract text from HTML pages and PDFs, deduplicate identical text across crawls while retaining all occurrences and source versions, add rich document-level annotations with propella~\citep{idahl2026propella}, and package a dataset of 122 million unique documents totaling 314 billion tokens.
We further characterize RefinedGAW in language, source, and content composition, and compare its German subset to large German reference datasets.
Automatic document-level annotations indicate a high concentration and substantial volume of academic research content in RefinedGAW's German subset~(Figure~\ref{fig:refinedgaw-academic}).
Our broader characterization also identifies substantial educational, math-heavy, and expert-oriented material, while marketing-heavy content is rare~(Figure~\ref{fig:refinedgaw-yields}).
Across HTML and PDF, 88\% of German RefinedGAW documents have no near-duplicate candidate match in the three evaluated reference datasets, supporting complementary coverage.
We publicly release a queryable metadata index including document-level annotations and source provenance, together with software for retrieving and processing source documents.
The annotations enable fine-grained corpus analysis and subset selection, the retained URLs, crawl occurrences, capture times, and source versions enable longitudinal study of institutional web content.

\section{Dataset Construction}
\label{sec:resource-construction}

RefinedGAW is based on 26 crawls from the German Academic Web~\citep{paris2020gaw}, covering October 2012 to June 2025.
The GAW snapshots are collected with the Heritrix crawler~\citep{mohr2004heritrix} from a seed list of roughly 150 domains covering the German academic institutions entitled to award doctorates, following links breadth-first within the seed domains and storing every capture as WARC records.
The crawler configuration excludes file types such as audio, video, and archives, hosts considered undesirable such as e-learning systems and repositories were retired during collection, and most crawls were stopped after roughly 100 million fetched pages.
Crawl-level analyses estimate that a single crawl reaches roughly half of the crawlable German academic web at the URL level~\citep{paris2026crawler,paris2026coverage}.

We process HTML pages and PDF documents in separate pipelines differing in text extraction and deduplication method, as detailed below and summarized in Table~\ref{tab:processing-summary}.
Each crawl is processed independently before exact-text deduplication across all crawls, separately for HTML and PDF.
Processing was performed on a single compute node with 80 cores and 4 TB RAM, unless noted otherwise.

\subsection{HTML Text Extraction and Deduplication}
\label{sec:html-processing}

First, we extract plain text from HTML pages with Trafilatura 2.0.0~\citep{barbaresi2021trafilatura}, using \texttt{output\_format="txt"}, \texttt{favor\_precision=True}, \texttt{include\_comments=False}, and \texttt{deduplicate=True}.
To exclude very short extraction outputs, we retain only documents yielding more than 300 characters.
We assign a content identifier to each retained document using the first 128 bits of its SHA-256 digest.
Failed extraction and the length threshold together remove 1.09 billion of the 1.88 billion HTML captures (58\%) leaving 785.8 million extracted documents.
This hash is computed over the UTF-8 encoded extractor output without any additional text normalization.
We then run global exact deduplication on this identifier across documents from all crawls, reducing 786 million extracted documents to 111 million unique texts, a reduction of 86\%.
For every unique text we retain metadata summaries of its crawls, capture frequencies, URLs, and time\-stamps, as well as links to textually different versions served from the same URL, to not discard any occurrence history.
This allows for longitudinal analysis and frequency-based sampling without storing redundant copies of the same text.
Of the 111 million deduplicated HTML texts, 32\% were observed in at least two crawls.
Separately, 31\% of text-crawl pairs aggregate multiple raw captures within a crawl, which can arise from repeated retrieval or different URLs yielding identical text.
We report the full recurrence distributions in Appendix~\ref{app:exact-text-recurrence}.

\subsection{PDF OCR and Deduplication}
\label{sec:pdf-processing}
Being more expensive than web page processing, our PDF processing pipeline begins with global deduplication to ensure the same file is rendered and OCRed only once.
Again, we retain provenance metadata for every duplicate.
This reduces the 84.9 million PDF captures to 15.0 million byte-distinct files requiring extraction, a reduction of 82\% that avoids running 69.8 million redundant OCR tasks.
We then rasterize each page at 300 dpi and perform text extraction with Unlimited OCR~\citep{yin2026unlimited}\footnote{Model revision \texttt{27a5997}.}, at a total cost of around 10k H100 hours.
We record page-level OCR outcomes and quality flags, and for each document concatenate all successfully processed pages in their original order.
Following our HTML processing pipeline, we again keep only texts of more than 300 characters, assign document identifiers based on the extracted text, and run another round of global exact deduplication based on this identifier.
Failed extraction and the length threshold together remove 2.0 million files, 14\% of those sent to OCR, leaving 13.0 million extracted texts.
This second deduplication stage catches identical text produced from byte-distinct PDF files, removing an additional 2.4 million documents and leaving 10.6 million unique PDF texts, a further reduction of 18\%.
PDFs recur more persistently than web pages: 58\% of unique texts appear in more than one crawl, and when a text appears in a crawl, that crawl captured it more than once 23\% of the time (Appendix~\ref{app:exact-text-recurrence}).

\begin{table}[t]
    \centering
    \small
    \setlength{\tabcolsep}{5.0pt}
    \begin{tabular}{lrrr}
        \toprule
        Pipeline stage & HTML & PDF & Total \\
        \midrule
        Captures & 1,879.7M & 84.9M & 1,964.6M \\
        Byte-distinct files & n/a & 15.0M & n/a \\
        Extracted texts & 785.8M & 13.0M & 798.8M \\
        Unique texts & 111.3M & 10.6M & 121.9M \\
        \midrule
        Share of captures & 5.9\% & 12.5\% & 6.2\% \\
        Tokens & 229.6B & 84.9B & 314.5B \\
        \bottomrule
    \end{tabular}
    \caption{Document counts at each stage of the HTML and PDF pipelines across the 26 GAW crawls.
    }
    \label{tab:processing-summary}
\end{table}

\subsection{Document-Level Annotation}
\label{sec:document-annotation}
We run language identification for all retained HTML and PDF documents using GlotLID v3~\citep{kargaran2023glotlid} once per document with the same procedure for HTML and PDF.
A PDF therefore receives a label for its concatenated text, not one label per page.
As model inputs we keep at most the first 40,000 characters, remove Markdown table rows and common formatting characters, and normalize whitespace.
Documents are labeled as unknown if they contain fewer than 50 UTF-8 bytes of alphabetic characters or have an alphabetic-to-non-whitespace byte ratio below 0.2.
We report on language distribution and threshold sensitivity in Appendix~\ref{app:language-distribution} and Table~\ref{tab:language-threshold-sensitivity}.

In addition, we annotate all retained documents with propella-1-4b~\citep{idahl2026propella}, a multi-property document annotation model.
It produces annotations in JSON format containing 18 properties, covering content integrity, content ratio, content length, content type, business sector, technical content, content quality, information density, educational value, reasoning indicators, audience level, commercial bias, time-sensitivity, content safety, PII presence, regional relevance, country relevance, and a high-level one-sentence description.
These annotations enable multi-dimensional dataset composition analysis and flexible subset selection.
Annotating the complete collection required approximately 2k H100 hours.

We further store per-document token counts obtained with the GPT-OSS tokenizer~\citep{openai2025gptoss} and the Gemma 3 tokenizer~\citep{gemmateam2025gemma3}.
All token counts reported in this paper are based on GPT-OSS.

\section{Dataset Characterization}
\label{sec:resource-analysis}

The HTML and PDF parts in RefinedGAW are kept separate throughout our analysis since their document lengths, extraction behavior, and content composition differ substantially.
We report both document- and token-centric statistics, the former showing how common a property is and the latter how much text it accounts for.

\subsection{Scale and Format Contributions}
\label{sec:resource-scale}
Table~\ref{tab:processing-summary} contains the release totals.
HTML and PDF formats contribute very unevenly.
Being much longer, PDFs account for 27\% of all tokens while making up only 9\% of documents, so document counts alone substantially understate how much content PDFs contribute to the dataset.
Appendix Figure~\ref{fig:refinedgaw-length-retention} reports the corresponding retention curves for the complete RefinedGAW HTML and PDF collections.
The median HTML document is 337 tokens and the median PDF text 1,292 tokens, but both distributions have long upper tails, and PDFs extend much further.
78 billion tokens, a quarter of the dataset, occur in documents of at least 131,072 tokens.
This data may support long-context training and adaptation, although document length alone does not establish discourse coherence or downstream utility.

\subsection{Language and Source Composition}
\label{sec:language-source}
The GAW has an institutional scope which means RefinedGAW is not a German-only corpus.
German is the most frequent top language label at 45\% of documents, but English documents are substantially longer, so the two are close in token share at 33\% and 35\% respectively.
The remainder is spread over other languages and low-confidence assignments.
Appendix~\ref{app:language-distribution} reports on the full document- and token-weighted distributions and their sensitivity to the language confidence thresholds.

Likewise, source concentration is also uneven, with the ten highest-yield HTML domains contributing 36\% of HTML tokens from only 5\% of HTML documents.
This means RefinedGAW should not be treated as a uniform sample of German academic institutions.
The host and crawl provenance in RefinedGAW lets users inspect or correct this imbalance when sampling.
Appendix Figures~\ref{fig:source-host-concentration} and~\ref{fig:temporal-contribution} report the leading hosts and how much each successive crawl adds.

\subsection{Annotation-Based Content Profile}
\label{sec:content-profile}

We characterize the content in RefinedGAW using the propella annotations introduced in Section~\ref{sec:document-annotation}, finding that both formats consist predominantly of academic content that the extraction pipelines recover cleanly.
Good or excellent content quality covers 75\% of HTML documents and 91\% of PDF texts.
Roughly two thirds of documents are labeled as academic research, an advanced or expert audience is indicated for 60\% of HTML documents and 71\% of PDF texts.
Heavy or pure marketing bias appears in only about 1\% of each.
The formats differ mainly in document character: PDFs are far more often analytical (32\% versus 7\%) and contain more content with moderate or high educational value (38\% versus 7\%), consistent with their greater length.
Figure~\ref{fig:refinedgaw-yields} presents the corresponding token yields, where documents labeled as academic research or scientific content account for 235 billion tokens, about three quarters of the entire dataset.

\section{Comparison with German Datasets}
\label{sec:comparison}
We define the German subset of RefinedGAW as all documents for which \texttt{deu\_Latn} is the top GlotLID label, and compare it with a set of German reference datasets, comprising the German subsets of FineWeb2~\citep{penedo2025fineweb2}, HPLT 3.0~\citep{oepen2026hplt} and FinePDFs~\citep{kydlicek2025finepdfs}, as well as the German Commons~\citep{gienapp2025germancommons} dataset.
We refer to the German subsets as RefinedGAW-DE, FineWeb2-DE, HPLT 3.0-DE, and FinePDFs-DE, respectively.
FineWeb2 and FinePDFs derive from Common Crawl~\citep{commoncrawl}, HPLT 3.0 additionally incorporates data from the Internet Archive. 
German Commons is collected from heterogeneous sources.
Note that each reference dataset applied its own language-identification pipeline and cutoff.
For example, to identify German text, FineWeb2 uses a document-level confidence threshold of 0.82, whereas FinePDFs uses approximately 0.4 on page-averaged scores.
In Appendix~\ref{app:language-distribution} we report how our results change for different minimum confidence thresholds.
We compare document lengths, textual overlap, and text content profiles based on propella-annotations.

\subsection{Document-Length Comparison}
\label{sec:length-comparison}

\begin{figure}[t]
    \centering
    \includegraphics[width=\columnwidth]{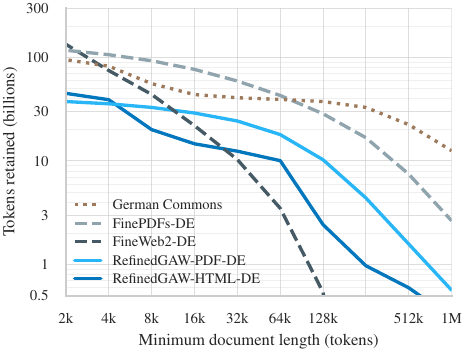}
    \caption{Token counts for German text datasets under varying minimum document-length thresholds.
    We report the total number of GPT-OSS tokens in documents containing at least that many tokens.
    Both axes use log-scales to show differences in the upper tails.
    }
    \label{fig:document-length-distributions}
\end{figure}

Figure~\ref{fig:document-length-distributions} shows how many tokens each dataset provides for varying minimum document-length thresholds.
RefinedGAW-HTML-DE has a slightly shorter median document than FineWeb2-DE (339 versus 422 tokens) and holds less token mass below roughly 32,000 tokens, reflecting its smaller size, but it overtakes FineWeb2-DE beyond that point and retains 2.4 billion tokens against 0.5 billion above 131,072 tokens.
For PDFs, FinePDFs-DE retains more at every threshold.
German Commons carries by far the heaviest long-document tail, with 36\% of its tokens in documents above 131,072 tokens, reflecting its book and legal sources.
RefinedGAW therefore contributes long web documents that broad-crawl datasets largely lose, but it is not the largest source of long-form German text.

\subsection{Overlap with Reference Datasets}

We measure overlap between dataset pairs directionally, from RefinedGAW to each reference dataset, using two indicators:
(1) An exact URL match, where a retained canonical or alternative URL occurs verbatim in the reference dataset, detects a shared web location.
(2) A fuzzy match, based on MinHash locality-sensitive hashing over distinct five-word shingles, as implemented in DataTrove~\citep{penedo2024datatrove}, additionally retrieves similar text served from different URLs.
German Commons is excluded from the URL comparison as it does not provide URLs.
The default MinHash configuration we use is tuned for recall and returns candidate pairs rather than verified duplicates.
Candidate coverage includes pairs below a chosen similarity threshold but might also miss pairs probabilistically, so the rates are an operational screen for possible overlap rather than a formal upper bound on near-duplicate prevalence.
Appendix~\ref{app:overlap-validation} details the configuration, validates candidate precision against exact Jaccard similarity, and reports the strong dependence of candidate rates on document length.

\begin{table}[t]
    \centering
    \small
    \setlength{\tabcolsep}{3.5pt}
    \begin{tabular}{@{}lrrrr@{}}
        \toprule
        & \multicolumn{2}{c}{HTML} & \multicolumn{2}{c}{PDF} \\
        \cmidrule(lr){2-3}\cmidrule(l){4-5}
        Reference dataset & URL & LSH & URL & LSH \\
        \midrule
        \multicolumn{5}{@{}l}{\textit{Complete release (RefinedGAW)}} \\
        FineWeb2-DE & 2.83\% & 5.08\% & 0.0\% & 1.62\% \\
        FinePDFs-DE & 0.0\% & 0.20\% & 9.74\% & 8.14\% \\
        German Commons & n/a & 0.03\% & n/a & 0.05\% \\
        \textbf{Any} & & \textbf{5.17\%} & & \textbf{9.26\%} \\
        \midrule
        \multicolumn{5}{@{}l}{\textit{German subset (RefinedGAW-DE)}} \\
        FineWeb2-DE & 6.17\% & 11.32\% & 0.0\% & 3.11\% \\
        FinePDFs-DE & 0.0\% & 0.44\% & 16.92\% & 15.57\% \\
        German Commons & n/a & 0.05\% & n/a & 0.10\% \\
        \textbf{Any} & & \textbf{11.47\%} & & \textbf{17.72\%} \\
        \bottomrule
    \end{tabular}
    \caption{Percentages of RefinedGAW-HTML and -PDF documents having an exact-URL match or a near-duplicate text candidate in German reference datasets.
    \emph{Any} is the union of the three reference datasets.
    URL overlap is unavailable for German Commons since it does not contain document URLs.
    }
    \label{tab:textual-overlap}
\end{table}

The overlap results in Table~\ref{tab:textual-overlap} support that RefinedGAW is a complementary resource.
Across the complete release, only 5\% of HTML and 9\% of PDF documents have a candidate match in any of the reference datasets.
Within RefinedGAW-DE, the corresponding candidate shares are 11\% for HTML and 18\% for PDF.
Pooling by document count across the two formats, 88\% of documents have no candidate match.
Overlap also concentrates within source formats.
RefinedGAW-HTML matches FineWeb2-DE and RefinedGAW-PDF matches FinePDFs-DE, matches across formats and German Commons are close to zero.

\subsection{Annotation-Based Content Comparison}

To compare content composition across datasets, we analyze the propella-based document annotations included in RefinedGAW.
For the reference datasets we rely on the publicly available annotations from~\citet{idahl2026propella}\footnote{\url{https://hf.co/datasets/openeurollm/propella-annotations}}, which were produced with the same model and inference configuration.

\begin{figure*}[t]
    \centering
    \includegraphics[width=\textwidth]{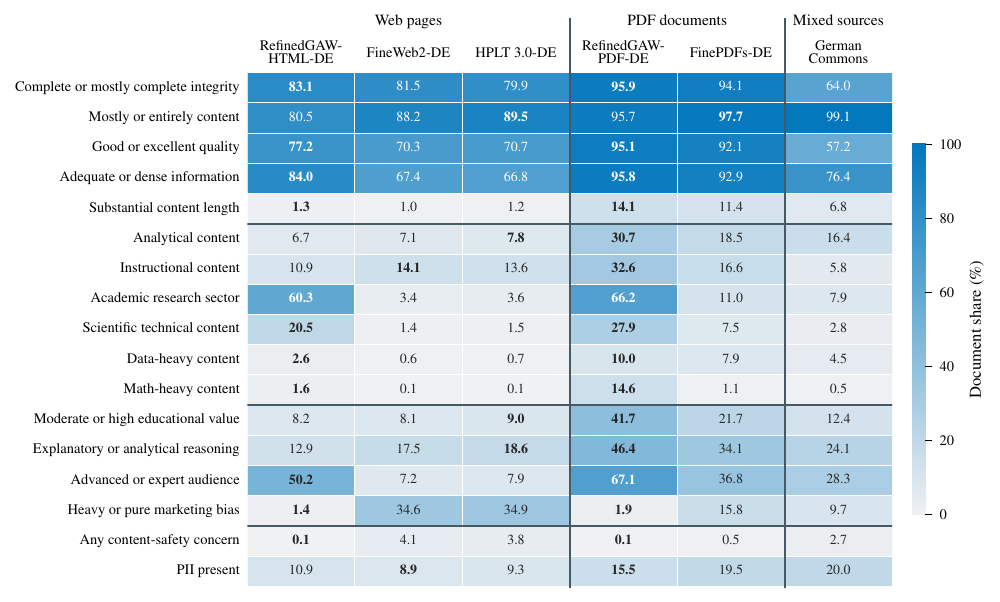}
    \caption{Prevalence of selected propella properties for German RefinedGAW subsets and German reference datasets.
    Each cell displays the percentage of documents falling in the stated category.
    Columns are grouped by source format, and horizontal rules separate four property groups: extraction quality, subject matter, purpose and audience, and safety and privacy.
    The largest value within each source-format group is marked in bold, except for marketing bias and the safety and privacy group, where lower is better.
    German Commons is a mixed-source dataset and belongs to neither group.
    Content-safety concerns comprise the labels \textit{mildly concerning}, \textit{NSFW}, \textit{harmful}, and \textit{illegal}.
    }
    \label{fig:propella-comparison}
\end{figure*}

In Figure~\ref{fig:propella-comparison} we plot what percentage of documents carry certain labels.
Academic research content is far more prevalent in both German RefinedGAW subsets.
It covers 60\% of RefinedGAW-HTML-DE documents but only 3\% of FineWeb2-DE and 4\% of HPLT 3.0-DE, and 66\% of RefinedGAW-PDF-DE documents against 11\% of FinePDFs-DE.
Scientific technical content, which rates technical sophistication rather than domain, follows the same ordering, at 21\% against 1\% for both web-page references and 28\% against 8\% for PDF.
Only 1\% and 2\% of the RefinedGAW subsets are labeled as heavy or pure marketing, against 35\% for both FineWeb2-DE and HPLT 3.0-DE, 16\% for FinePDFs-DE, and 10\% for German Commons.
Content-safety concerns are rare in both RefinedGAW subsets, at 0.1\%, against 4\% for the web-page references and 0.5\% for FinePDFs-DE.
Since we do not apply any safety filters, this most plausibly reflects the institutional scope of the underlying crawl.
The extraction indicators are mixed.
Good or excellent content quality is higher for both formats, at 77\% against 70 to 71\% for HTML and 95\% against 92\% for PDF, and information density is markedly higher, at 84\% against 67\%.
Content integrity is comparable within each source format, at 80 to 83\% for the web-page datasets and 94 to 96\% for the PDF datasets, with German Commons the exception at 64\%.

Document prevalence alone does not indicate how much text each dataset supplies, so we also compare absolute token counts.
Academic research accounts for 83\% of RefinedGAW-HTML-DE tokens and 84\% of RefinedGAW-PDF-DE tokens, well above the corresponding document shares, which indicates that these documents are longer than average.
When the format-specific counts are summed, the German RefinedGAW subsets contain 87.6 billion academic-research tokens within 104.8 billion tokens overall, compared with 64.9 billion within 516.0 billion tokens across FineWeb2-DE and FinePDFs-DE.
Under the respective corpus selection procedures, RefinedGAW contains 35\% more tokens in documents labeled as academic research than these two reference datasets combined, despite containing only about one fifth as many German-labeled tokens overall.
Absolute yields depend on language selection and deduplication.
Appendix~\ref{app:language-distribution} reports the sensitivity of German token yields to language-confidence thresholds.
RefinedGAW-HTML-DE holds 64.5 billion tokens against FineWeb2-DE's 381.5 billion, yet contributes more academic-research tokens, 53.6 against 22.3 billion (Figure~\ref{fig:refinedgaw-academic}), so its concentration outweighs the six-fold difference in size.
Commercially biased and unsafe content is rare.
RefinedGAW-HTML-DE contains only 0.5 billion marketing-heavy tokens against FineWeb2-DE's 143.7 billion, and 0.2 billion tokens with content-safety concerns against 32.8 billion.
RefinedGAW-PDF-DE is smaller than FinePDFs-DE and contributes fewer academic-research tokens, 34.0 against 42.6 billion, with math-heavy content the exception at 8.7 against 5.0 billion.
Appendix Figure~\ref{fig:propella-token-yields} reports these token yields in full, Appendix~\ref{app:language-distribution} reports how the comparison shifts when imposing a minimum confidence score during language identification.
Token counts are not available for HPLT 3.0-DE, which we did not tokenize, so the token comparison covers the remaining five corpora.

However, the profile does not uniformly favor RefinedGAW.
RefinedGAW-HTML-DE has a lower share of documents labeled as mostly or entirely content, 80\% against 88\% and 90\%, which may reflect the navigation-heavy layout of institutional pages.
These labels indicate at least 70\% substantive content rather than navigation, interface elements, or structural markup under the propella rubric.
Reasoning indicators fall below the web-page references, 13\% against 17\% and 19\%, while the share of educational documents is even with FineWeb2-DE at 8\% and slightly below HPLT 3.0-DE at 9\%.
Personally identifiable information is also flagged more often in RefinedGAW-HTML-DE than in the broad-web references, 11\% against 9\%, which is consistent with institutional pages such as staff directories and contact listings; for PDFs however the ordering reverses.
Several of these differences are consistent with the composition of institutional web pages, although the content ratio is also sensitive to extractor configuration and we do not separate the two effects.
RefinedGAW is therefore best understood as complementary to the broad-web references rather than as a higher-quality substitute.

\section{Release, Selection, and Rehydration}
\label{sec:resource-release}

\subsection{Metadata Index and Document Selection}
\label{sec:index-selection}

We release RefinedGAW as a metadata index rather than as archived text, containing document identifiers, source URLs, crawl and version provenance, length statistics, propella annotations, and quality scores.
The reason is not specific to this project.
Under the European text and data mining exceptions, the conditions for retaining and redistributing a curated corpus are considerably less settled than those for mining it, and European research consortia consistently report that this uncertainty prevents them from releasing corpora they have already built~\citep{drabczyk2026copyright}.
The GAW archive publishes URLs and limited metadata, and our release follows both their practice and the precedent set by LAION-5B~\citep{schuhmann2022laion5b}, which likewise releases URLs and metadata while leaving retrieval of the referenced content to individual users.
The GAW crawler observes \texttt{robots.txt} at crawl time, but this is a weaker safeguard for the earlier crawls than for the later ones, since the series begins in 2012 and machine-readable reservations directed at text and data mining did not exist for most of its span.
Captures collected before such reservations were available cannot be filtered retroactively, so the rehydration tool instead applies current \texttt{robots.txt} rules at retrieval time.
The index is licensed under the Creative Commons Attribution 4.0 International License (CC BY 4.0), which covers the index and derived metadata but not the source documents referenced by its URLs.
Users can query the annotations on the Hugging Face Hub, define subsets, and retrieve only the corresponding source documents.
For convenience, a heuristic \texttt{quality\_score}, described in Appendix~\ref{app:quality-score}, provides a default document ordering.

We release an accompanying rehydration tool\footnote{\url{https://github.com/maxidl/refinedgaw-rehydrate}} (Apache-2.0), that can be used to retrieve currently public resources and run the text extraction pipeline.
It also compares the resulting text hashes with the dataset identifiers, making exact recovery directly observable, although for PDFs a mismatch can also arise from OCR nondeterminism (Section~\ref{sec:rehydration}).
Subject to applicable law, we may grant named researchers controlled, time-limited access to the full-text version of RefinedGAW for non-commercial replication studies, without onward distribution.

\subsection{Present-Day Rehydration}
\label{sec:rehydration}

Because this release design depends on live source URLs, we run a rehydration experiment to measure how many documents can still be recovered from the public web.
For each format we randomly select 5,000 records from the top 10\% of the quality ranking and 5,000 from the remaining 90\% with a fixed seed, oversampling the top group so that its recovery rate can be compared with the rest.
Using rehydration tool version 0.1.3 on August 3, 2026, we attempted to fetch the public source URLs in deterministic order under conservative \texttt{robots.txt}, rate, retry, and size limits.

Table~\ref{tab:url-reachability}(a) reports the share of sampled records recovered under four successively stricter criteria, from downloading the source to reproducing the archived text exactly.
Every rate uses the complete sample as its denominator rather than only records passing the preceding stage.
We describe the exact criteria in Appendix~\ref{app:rehydration-diagnostics}.

\begin{table}[h]
    \centering
    \small
    \setlength{\tabcolsep}{2.2pt}
    \begin{tabular*}{\dimexpr\columnwidth-2pt\relax}{@{\extracolsep{\fill}}llrrrr@{}}
        \multicolumn{6}{@{}l}{\textit{(a) Outcomes by selection group}} \\
        \toprule
        Format & Sample & Down. & Extract. & Jacc. $\geq .8$ & Exact \\
        \midrule
        HTML & Top 10\% & 44.8\% & 41.2\% & 21.0\% & 11.7\% \\
        HTML & Other 90\% & 34.6\% & 27.5\% & 8.1\% & 4.3\% \\
        PDF & Top 10\% & 50.7\% & 50.2\% & 35.7\% & 7.2\% \\
        PDF & Other 90\% & 39.3\% & 39.0\% & 23.6\% & 8.3\% \\
        \bottomrule
    \end{tabular*}
    \vspace{6pt}
    \setlength{\tabcolsep}{3.5pt}
    \begin{tabular*}{\dimexpr\columnwidth-2pt\relax}{@{\extracolsep{\fill}}lccc@{}}
        \multicolumn{4}{@{}l}{\textit{(b) Downloaded records by crawl date}} \\
        \toprule
        Format & 2013-02 & All crawls & 2025-06 \\
        \midrule
        HTML & 19.1\% & 35.7\% & 75.2\% \\
        PDF & 17.7\% & 40.4\% & 82.6\% \\
        \bottomrule
    \end{tabular*}
    \caption{Present-day rehydration success rates by quality-ranking group and crawl recency.
    (a) reports the percentage of all 5,000 sampled records per group reaching each cumulative stage: source downloaded (Down.), text extracted (Extract.), Jaccard similarity of at least 0.8 over normalized five-word shingles, and exact equality with the archived text.
    (b) reports source-download rates for a selected early crawl shared by both formats, the latest crawl, and all crawls combined; rates weight the Top 10\% and Other 90\% samples by 0.1 and 0.9 to represent their population shares.}
    \label{tab:url-reachability}
\end{table}

Rehydration recovers a minority of the index: 36\% of HTML records and 40\% of PDF records can still be downloaded today.
Nearly every downloaded PDF still yields text, against four in five for HTML, and 56\% of downloaded PDFs are byte-identical to the archived file, or 23\% of all PDF records.
Exact reproduction of the stored text is nonetheless rare for both formats, at 8\% of PDF and 5\% of HTML records.
For HTML this is consistent with live pages changing after capture, whereas for PDF the source file is often unchanged and the extraction is not reproducible: only 36\% of byte-identical files yield the stored text.
Missing resources, current \texttt{robots.txt} exclusions, and DNS failures account for most unsuccessful downloads in both formats.
Table~\ref{tab:url-reachability}(a) reports the full outcome ladder for each selection group, and bootstrap 95\% confidence intervals span at most 2.5 percentage points throughout.

\paragraph{Crawl recency.}
Recovery depends strongly on source age: HTML availability rises from 19\% for February 2013 to 75\% for June 2025, and PDF from 18\% to 83\% (Table~\ref{tab:url-reachability}b).
Part of the Top 10\% advantage is therefore a date-composition effect, and Appendix~\ref{app:rehydration-diagnostics} reports the year-specific estimates and the standardized comparison.

\section{Related Work}
\label{sec:related-work}

\subsection{Longitudinal Academic Web Archives}

Common Crawl supplies the principal open broad-crawl infrastructure for modern web corpora~\citep{commoncrawl}.
National and thematic web archives pursue a different goal, preserving longitudinal collections that are usually organized for historical access rather than distributable NLP data.
Within the academic web, \citet{hale2014ukwebspace} demonstrate the research value of an institutionally bounded archive through a longitudinal analysis of the British \texttt{.ac.uk} domain, but do not release a reusable document corpus.
GAW likewise supports longitudinal study~\citep{paris2020gaw}, and subsequent work uses it to examine how crawl configuration affects exhaustiveness~\citep{paris2026crawler,paris2026coverage}.
The selected pages studied by \citet{younes2022datasets} further show the value of GAW content, but required task-specific access and preprocessing.
RefinedGAW changes the mode of access by integrating extracted HTML and PDF text, global exact deduplication, URL and capture provenance, and document-level annotations into a general NLP resource.

\subsection{German and Multilingual Web Corpora}

Large multilingual web corpora offer broad German coverage through several collection strategies.
FineWeb2 processes almost 100 Common Crawl snapshots with language-adaptive filtering and global per-language deduplication~\citep{penedo2025fineweb2}, whereas HPLT 3.0 combines Common Crawl with Internet Archive crawls and enriches documents with quality, register, and personally identifiable information annotations~\citep{oepen2026hplt}.
Aleph-Alpha-GermanWeb applies German-specific heuristic and model-based curation to Common Crawl, FineWeb2, and synthetic text~\citep{burns2026germanweb}.
German Commons instead aggregates 41 explicitly licensed sources across seven domains and applies source-specific processing and deduplication~\citep{gienapp2025germancommons}.
KletterMix expands German pretraining resources through translation of an English corpus while preserving document boundaries and source metadata~\citep{kraus2026klettermix}.
What distinguishes RefinedGAW is not another filtering recipe for the same crawl base, but access to the repeated captures and source history of an independently operated academic crawl.

\subsection{Scientific-Document and PDF Corpora}

Scientific-document corpora provide a second point of comparison, but cover a narrower part of the academic information landscape.
French Science Commons aggregates articles and theses from repositories and scholarly indexes, preserving document-level source, license, and bibliographic metadata~\citep{pleias2026frenchscience}.
Because GAW follows institutional websites, RefinedGAW also includes teaching materials, administrative documents, project pages, staff profiles, and news.
This broader institutional record complements publication-centered collections.

FinePDFs is the closest large-scale comparator for RefinedGAW's PDF component, extracting and filtering PDFs collected across Common Crawl snapshots in more than 1,000 languages~\citep{kydlicek2025finepdfs}.
The comparison separates RefinedGAW's focused institutional PDF collection from PDFs encountered by a broad web crawl.

\section{Conclusion}
\label{sec:conclusion}

We processed 26 existing GAW crawls into 122 million unique, annotated HTML and PDF texts totaling 314 billion tokens while retaining crawl-occurrence and source-version provenance.
Of the German subset, 87.6 billion tokens come from documents labeled as academic research, compared with 64.9 billion across FineWeb2-DE and FinePDFs-DE, whose combined German subset is almost five times larger.
The overlap analysis provides a second indication of complementarity: 88\% of German RefinedGAW documents have no candidate match in any of the three evaluated German reference datasets.
Together, these results show that total corpus size is a poor proxy for domain-specific yield and that existing focused archives can contribute material beyond much larger broad-web collections.
The annotations and provenance support content-aware selection and longitudinal analysis, while the public metadata index and rehydration software make present-day source recovery observable without publicly redistributing archived text.
Evaluating selected subsets in language-model training and studying institutional web change with the retained provenance remain future work.

\section*{Limitations}
\label{sec:limitations}

RefinedGAW inherits GAW's institutional and host coverage.
German-language academic writing outside the selected hosts is absent, and GAW does not capture every page within them.
A single crawl covers an estimated half of the crawlable German academic web at the URL level~\citep{paris2026crawler,paris2026coverage}.
Link structure, crawl limits, host exclusions, failed requests, and configuration changes alter the captured set.
For longitudinal analysis, a difference between snapshots is consequently ambiguous: it may reflect collection conditions or a change on the website itself~\citep{paris2020gaw}.

HTML extraction and neural PDF OCR can omit, reorder, or corrupt text, while the minimum-length filter removes documents of 300 characters or fewer that may still be useful.
Exact-text deduplication is performed separately for HTML and PDF and leaves fuzzy near-duplicates in place, so the combined token counts reported across formats are sums rather than the size of a cross-format-deduplicated union.
Repeated material across formats may therefore inflate the pooled counts for both RefinedGAW and the combined FineWeb2 and FinePDFs reference collections, and we do not measure how alternative extraction or deduplication choices would affect corpus composition or downstream use.

Two parts of the analysis lack collection-specific human validation.
Propella supplies the content labels, but we did not compare its predictions on RefinedGAW against human judgments.
Applying the same model to every corpus supports direct comparison while carrying any systematic model error into every result.
The heuristic quality score likewise has no calibration against human judgments or downstream outcomes.
Property token yields also use document-level attribution: the complete document is counted whenever it carries the corresponding label, although individual passages may not exhibit that property.
Language selection changes the absolute German token comparison.
RefinedGAW includes every document whose top GlotLID label is \texttt{deu\_Latn}, with no minimum confidence score; the reference corpora use their own methods and cutoffs.
The comparison populations consequently differ by construction, with short, multilingual, and OCR-corrupted documents especially susceptible to misclassification.
Appendix~\ref{app:language-distribution} quantifies the sensitivity of the resulting token counts.
HPLT 3.0-DE is absent from this token comparison because we did not tokenize it.

Our overlap analysis is directional and covers three selected reference datasets.
The URL component compares strings verbatim, without normalization.
The 88\% result comes from MinHash candidate retrieval.
Candidate samples contain pairs below 0.8 exact Jaccard similarity, so candidate rates can exceed verified near-duplicate prevalence; probabilistic retrieval, input truncation, and shingle representation can also hide genuine textual relationships.
The reported 88\% is therefore the share for which our recall-oriented procedure returned no candidate.
Establishing absence from all existing corpora would require a broader, verified comparison.

The public release contains metadata and no archived text, so it cannot reproduce a fixed full-text snapshot.
Rehydration depends on continued source availability, current \texttt{robots.txt} rules, extractor behavior, and nondeterministic PDF OCR.
Recent crawls and durable URLs are overrepresented in recovered data.
We do not evaluate selected subsets through language-model training or use the retained provenance in a substantive study of institutional change.
Accordingly, this paper makes no empirical claim about language-model value or longitudinal findings.

\section*{Ethical Considerations}
\label{sec:ethical-considerations}

RefinedGAW derives from third-party pages and documents.
Their authors did not necessarily consent to redistribution, annotation, or language-model training.
Public availability and compliance with \texttt{robots.txt} establish neither author consent nor an unrestricted license for downstream use.
For curated-corpus efforts in the EU, the legal basis for public full-text redistribution is unsettled~\citep{drabczyk2026copyright}.
We therefore release source URLs and derived metadata, with no archived text.
The CC BY 4.0 license applies only to the index and metadata (Section~\ref{sec:index-selection}).
The index can reduce processing costs substantially.
Users can query the existing annotations and provenance to select a subset and retrieve only its sources, so retrieval and extraction costs scale with the selected subset.
Changed source text may require renewed annotation.
Aggregate source-download rates are 36\% for HTML and 40\% for PDF, with substantial variation by source age and quality-ranking group (Section~\ref{sec:rehydration}).
We report these rates as an estimate of what the present framework costs a single curated corpus, and note that a statutory right to share research datasets for verification has been proposed in response to this pattern~\citep{drabczyk2026copyright}.
The rehydration software retrieves only currently public sources under current \texttt{robots.txt} rules and conservative request limits.
Any separately approved full-text access would be controlled, time-limited, restricted to named researchers conducting non-commercial replication, and prohibit onward distribution.
A contact route permits metadata correction or removal.
These safeguards reduce redistribution risk, but users must still assess individual sources under applicable law, licenses, and source terms.

Institutional websites routinely contain personal information.
Staff and student directories, author pages, and event notices identify people; historical captures can retain names, affiliations, or contact details after the live page changes.
URLs and derived metadata can themselves reveal an identity or sensitive context.
Propella's binary PII label can screen material but provides no anonymization and may miss instances.
Users should limit retrieval to what their research requires, conduct a context-specific privacy review, and validate automated filters on German academic material.
Human review is necessary when errors could affect identifiable people.
RefinedGAW is not intended for longitudinal profiling, surveillance, or consequential assessment of people or institutions.

GAW's coverage is institutionally skewed.
It centers doctorate-granting institutions and their public websites, leaving academic work outside the selected hosts and material missed through crawl policy or access conditions unrepresented.
Models trained on RefinedGAW may reproduce these gaps along with social and institutional biases in the source text.
Low estimated rates of unsafe or commercial content do not guarantee their absence, and systematic annotation errors may also enter the quality score.
The annotations and score are unsuitable for evaluating people, institutions, or documents.
They must not be used to allocate opportunities or support consequential decisions.

Retrieved documents may contain copyrighted passages, personal information, examination material, or benchmark examples.
A language model can memorize or reproduce this material.
Before training, users should review source permissions and intended use, apply privacy and safety controls to the selected subset, and perform appropriate benchmark decontamination.
Inclusion in RefinedGAW does not certify a document as suitable training data.

Neural OCR required approximately 10k H100 hours (Section~\ref{sec:pdf-processing}), and document annotation with propella required approximately 2k H100 hours.
Releasing annotations avoids repeating their inference cost, although rehydrating PDF text still requires document retrieval and extraction.

\section*{Acknowledgments}

This research is supported by the OpenEuroLLM project, co-funded by the Digital Europe Programme under GA no. 101195233.
This project is supported by the LLMs4EU project, co-funded by the Digital Europe Programme under GA no. 101198470.
This project is supported by the German Federal Ministry for Economic Affairs and Energy (BMWE) through EU-SAI/SOOFI: Sovereign Open Source Foundation Models for European Intelligence (grant number 13IPC040J).

\bibliography{references}

\appendix

\section{Token Mass Above Document-Length Thresholds}
\label{app:length-retention}

Document medians do not show where token mass lies within the strongly right-skewed length distributions.
At a minimum length of 2,048 tokens, the HTML and PDF collections retain 185.5 billion tokens~(81\% of HTML tokens) and 79.9 billion tokens~(94\% of PDF tokens), respectively.
Even at 131,072 tokens, they retain 57.7 billion and 20.7 billion tokens, together accounting for 25\% of all RefinedGAW tokens.
These counts measure extracted-text length and do not establish that the retained documents form coherent long-context sequences.

\begin{figure}[t]
    \centering
    \includegraphics[width=\columnwidth]{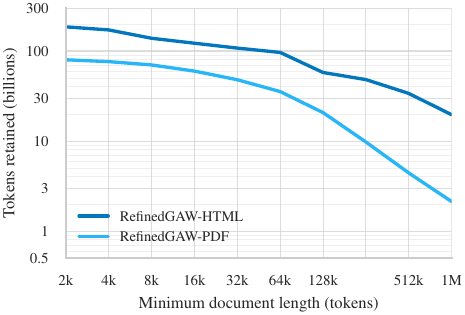}
    \caption{Token mass above minimum document-length thresholds in the complete RefinedGAW collections.
    Each point sums GPT-OSS tokens over documents containing at least the indicated number of tokens~\citep{openai2025gptoss}.
    Both axes use logarithmic scales to show the upper tails of the distributions.
    }
    \label{fig:refinedgaw-length-retention}
\end{figure}

\section{Exact-Text Recurrence}
\label{app:exact-text-recurrence}

\begin{figure}[t]
    \centering
    \includegraphics[width=\columnwidth]{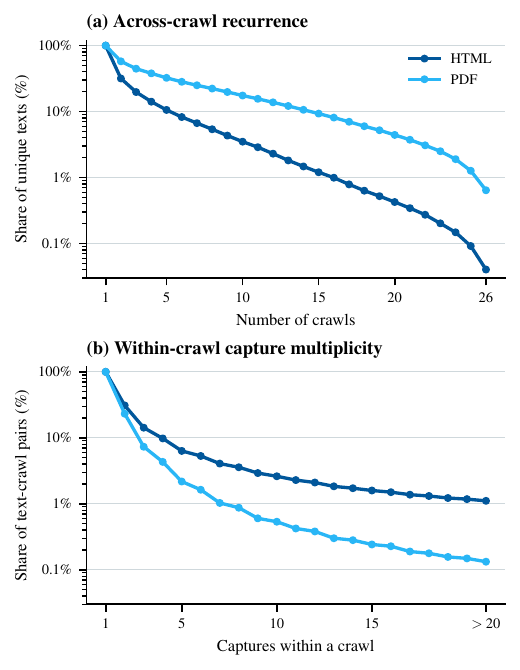}
    \caption{Exact-text recurrence across and capture multiplicity within GAW crawls.
    Panel (a) uses unique texts as its denominator and reports the share observed in at least the indicated number of the 26 crawls.
    Panel (b) uses text-crawl pairs as its denominator and reports the share with at least the indicated number of captured occurrences within that crawl; the final point pools all counts above 20.
    Curves are computed separately for RefinedGAW-HTML and RefinedGAW-PDF, and the vertical axes use logarithmic scales.
    }
    \label{fig:document-duplication}
\end{figure}

Exact-text deduplication stores one copy of each retained text while preserving its crawl and capture provenance.
Figure~\ref{fig:document-duplication} distinguishes recurrence across snapshots from repeated capture within a single snapshot.
Across snapshots, 32\% of HTML texts and 58\% of PDF texts occur in at least two of the 26 crawls, while 44,859 HTML texts and 68,001 PDF texts occur in all 26.
Within a crawl, 31\% of HTML text-crawl pairs and 23\% of PDF text-crawl pairs contain more than one capture.
PDF texts are therefore more persistent across snapshots, whereas repeated capture within a snapshot is more prevalent for HTML.

\section{Language Identification and Subset Sensitivity}
\label{app:language-distribution}

\subsection{Predicted Language Composition}

GAW follows institutions rather than a language boundary, so RefinedGAW is multilingual and language labels describe rather than filter the release.
Under the threshold-free top-label assignment, German and English together account for 76\% of HTML documents but 65\% of HTML tokens, and for 77\% of PDF documents and 79\% of PDF tokens.
The difference between document and token shares is especially pronounced for HTML, where small groups of long documents can contribute substantial token mass.
For example, texts predicted as Latin account for 0.3\% of HTML documents but 4.3\% of HTML tokens.
These categories are model predictions rather than human-verified language labels, and unusual assignments can reflect multilingual, formula-heavy, or extraction-degraded text.

\begin{figure*}[t]
    \centering
    \includegraphics[width=\textwidth]{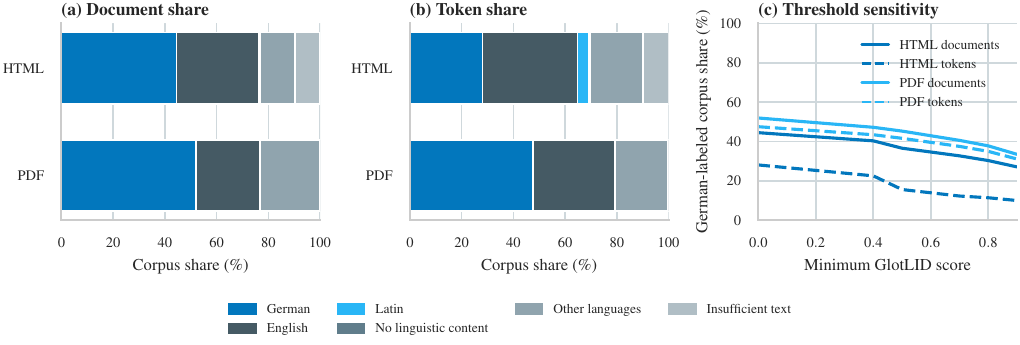}
    \caption{Predicted document-level language composition of RefinedGAW.
    GlotLID v3 assigns one label from at most the first 40,000 characters of each document; PDF pages are concatenated before inference.
    Panels (a) and (b) give document and GPT-OSS-token shares for the top prediction without a minimum score.
    Insufficient text marks documents that fail the alphabetic-text precheck, no linguistic content combines GlotLID's corresponding labels, and all other predicted languages are aggregated.
    Latin is displayed separately because it is the largest prediction outside German and English by token share and is concentrated in a small number of long HTML documents.
    Panel (c) traces the document and token shares assigned \texttt{deu\_Latn} as the confidence threshold increases.
    Token shares use document-level GPT-OSS counts~\citep{openai2025gptoss}.}
    \label{fig:language-distribution}
\end{figure*}
\FloatBarrier

\subsection{Sensitivity of the German Subset}

Section~\ref{sec:comparison} defines RefinedGAW-DE by the top GlotLID label without a minimum confidence score.
This threshold-free definition follows the release design, which does not filter documents by language confidence, and avoids choosing a cutoff to mirror any single comparator.
The reference datasets use different language-selection procedures: FinePDFs applies an approximate cutoff of 0.405 to page-averaged scores~\citep{kydlicek2025finepdfs}, whereas FineWeb2 applies a document-level cutoff of 0.821~\citep{penedo2025fineweb2}.
Applying either value to RefinedGAW would therefore not reproduce the corresponding reference population.
Table~\ref{tab:language-threshold-sensitivity} instead treats a range of minimum GlotLID scores as a sensitivity analysis.

The threshold is much more consequential for HTML than for PDF.
At 0.4, HTML retains 51.8 billion tokens, 80\% of its top-label German token mass, and 41.8 billion academic-or-scientific tokens, 78\% of the corresponding top-label yield.
PDF retains 36.9 billion tokens and 31.9 billion academic-or-scientific tokens at the same cutoff, both 91\% of their top-label values.
Between the top-label definition and a 0.9 cutoff, the HTML academic-or-scientific share falls from 83.5\% to 65.9\%, while its moderate-or-high educational share rises from 14.3\% to 33.5\%.
For PDF, the academic-or-scientific, advanced-audience, and no-commercial-bias shares each change by less than two percentage points, although the educational share rises from 69.9\% to 79.9\%.
The divergence between document and token curves in Figure~\ref{fig:language-distribution}c further shows that stricter thresholds disproportionately remove token-heavy HTML documents.
GlotLID confidence is therefore associated with document length and content composition rather than acting as a content-neutral precision control.

The overlap result changes for the same reason.
The format-matched candidate rate for HTML rises from 11.32\% under the top-label definition to 12.47\% at 0.4 and 16.53\% at 0.8, while the PDF rate rises from 15.57\% to 17.03\% at 0.4 because both the composition and denominator of the query population change.
At these format-relevant cutoffs, approximately 83\% of each RefinedGAW subset has no candidate in its format-matched reference dataset, so the evidence for complementary coverage remains while its precise magnitude depends on language selection.

Absolute HTML yield is more sensitive.
FineWeb2-DE contains 27.4 billion academic-or-scientific tokens, whereas RefinedGAW-HTML-DE contains 53.8 billion under the top-label definition and 41.8 billion at 0.4, but 26.1 billion at 0.5.
The RefinedGAW HTML yield therefore exceeds FineWeb2-DE under the first two definitions but not under stricter cutoffs; these rows quantify sensitivity rather than reconstructing FineWeb2's differently produced language subset.

\begin{table*}[t]
    \centering
    \small
    \begin{tabular}{@{}llrrrrr@{}}
        \toprule
        Format & German criterion & German & Academic or & Moderate/high & Advanced/expert & No commercial \\
        & & tokens & scientific & education & audience & bias \\
        \midrule
        HTML & Top label & 64.5 & 53.8 (83.5\%) & 9.2 (14.3\%) & 49.8 (77.3\%) & 60.1 (93.2\%) \\
        HTML & Score $\geq 0.4$ & 51.8 & 41.8 (80.7\%) & 9.0 (17.4\%) & 38.2 (73.8\%) & 47.5 (91.9\%) \\
        HTML & Score $\geq 0.5$ & 35.8 & 26.1 (72.8\%) & 8.9 (24.8\%) & 22.6 (63.0\%) & 31.6 (88.4\%) \\
        HTML & Score $\geq 0.7$ & 28.2 & 19.0 (67.4\%) & 8.5 (30.0\%) & 15.7 (55.7\%) & 24.2 (85.9\%) \\
        HTML & Score $\geq 0.8$ & 26.2 & 17.4 (66.4\%) & 8.1 (31.0\%) & 14.2 (54.2\%) & 22.3 (85.2\%) \\
        HTML & Score $\geq 0.9$ & 23.0 & 15.1 (65.9\%) & 7.7 (33.5\%) & 12.1 (52.8\%) & 19.3 (83.9\%) \\
        \midrule
        PDF & Top label & 40.4 & 35.0 (86.8\%) & 28.2 (69.9\%) & 36.0 (89.1\%) & 38.4 (95.2\%) \\
        PDF & Score $\geq 0.4$ & 36.9 & 31.9 (86.4\%) & 26.9 (73.0\%) & 32.9 (89.1\%) & 35.0 (94.9\%) \\
        PDF & Score $\geq 0.5$ & 35.3 & 30.5 (86.3\%) & 26.3 (74.5\%) & 31.5 (89.1\%) & 33.5 (94.8\%) \\
        PDF & Score $\geq 0.7$ & 31.9 & 27.4 (86.0\%) & 24.6 (77.4\%) & 28.4 (89.1\%) & 30.1 (94.5\%) \\
        PDF & Score $\geq 0.8$ & 29.7 & 25.5 (85.7\%) & 23.4 (78.7\%) & 26.5 (89.1\%) & 28.0 (94.4\%) \\
        PDF & Score $\geq 0.9$ & 26.4 & 22.5 (85.3\%) & 21.1 (79.9\%) & 23.5 (89.0\%) & 24.8 (94.0\%) \\
        \bottomrule
    \end{tabular}
    \caption{Sensitivity of selected token yields to the minimum GlotLID score among documents whose top label is \texttt{deu\_Latn}.
    Values are billions of GPT-OSS tokens, with the percentage of retained German token mass assigned to each outcome in parentheses.
    ``Top label'' applies no minimum confidence score.
    Academic or scientific is the union of documents labeled as academic research or scientific technical content.
    Outcomes can overlap and must not be summed.
    }
    \label{tab:language-threshold-sensitivity}
\end{table*}
\FloatBarrier

\subsection{German- and English-Labeled Content}

The English-labeled subset contributes 111.0 billion tokens across both formats, slightly more than the 104.8 billion in the German-labeled subset.
Table~\ref{tab:german-english-content} separates these populations by format and content property.
For HTML, the English subset is larger and contains substantially more educational and advanced-audience text, although academic-or-scientific content has a higher share and absolute yield in the German subset.
For PDF, the English subset is smaller but has a higher share in every displayed outcome.

\begin{table*}[t]
    \centering
    \small
    \begin{tabular}{@{}llrrrrr@{}}
        \toprule
        Format & Top language & Retained & Academic or & Moderate/high & Advanced/expert & No commercial \\
        & label & tokens & scientific & education & audience & bias \\
        \midrule
        HTML & German & 64.5 & 53.8 (83.5\%) & 9.2 (14.3\%) & 49.8 (77.3\%) & 60.1 (93.2\%) \\
        HTML & English & 84.2 & 46.4 (55.1\%) & 36.5 (43.4\%) & 79.8 (94.7\%) & 82.5 (98.0\%) \\
        \midrule
        PDF & German & 40.4 & 35.0 (86.8\%) & 28.2 (69.9\%) & 36.0 (89.1\%) & 38.4 (95.2\%) \\
        PDF & English & 26.8 & 24.6 (91.8\%) & 23.5 (87.6\%) & 26.2 (97.9\%) & 26.0 (97.3\%) \\
        \bottomrule
    \end{tabular}
    \caption{Content yields for the German- and English-labeled RefinedGAW subsets, defined by the top GlotLID label without a minimum score.
    Values are billions of GPT-OSS tokens, with the percentage of retained token mass assigned to each outcome in parentheses.
    Outcomes are multi-label and must not be summed.
    }
    \label{tab:german-english-content}
\end{table*}
\FloatBarrier

\section{Token Yields by Content Property}
\label{app:token-yields}

\begin{figure*}[t]
    \centering
    \includegraphics[width=\textwidth]{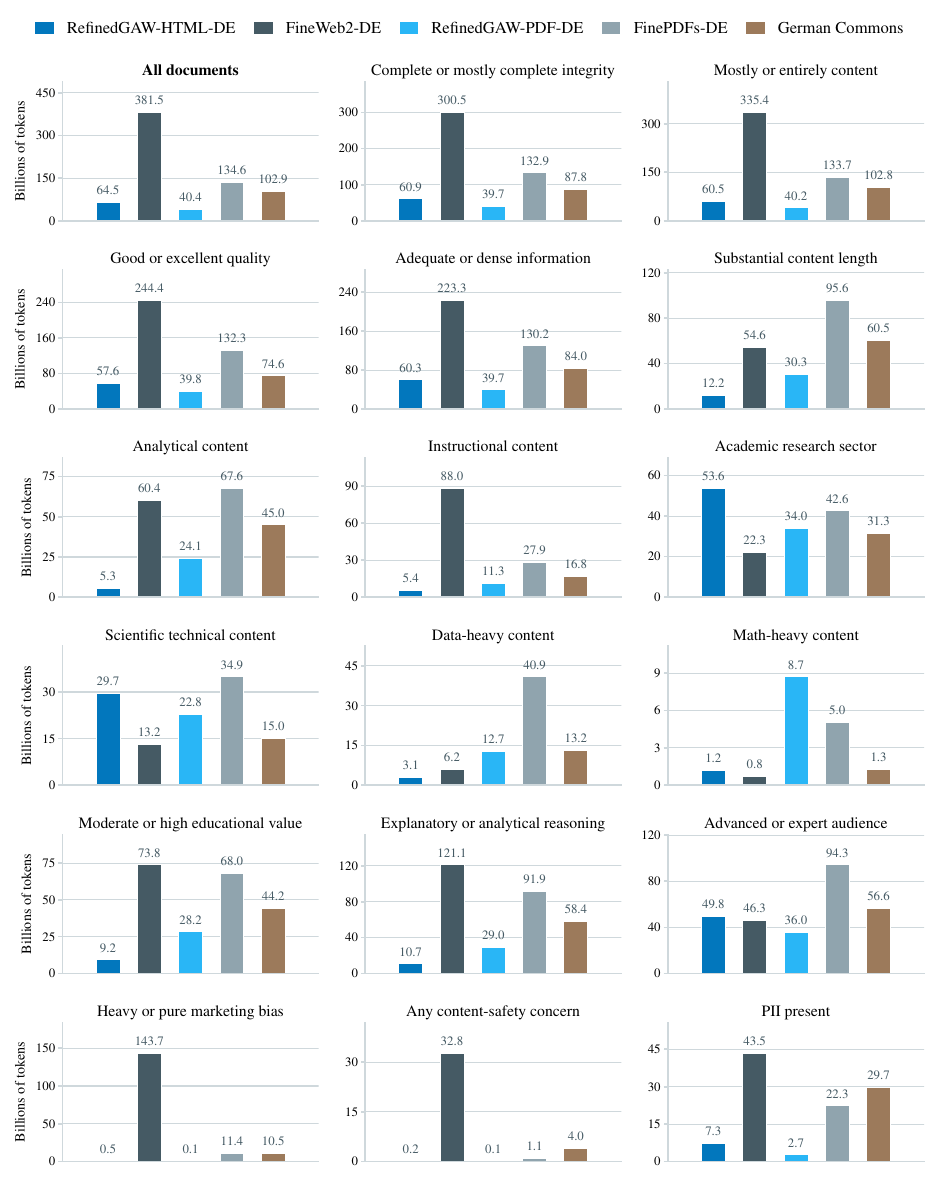}
    \caption{Absolute token yields for selected propella outcomes under the German-language populations used in Figure~\ref{fig:propella-comparison}.
    The RefinedGAW bars retain documents whose top GlotLID label is \texttt{deu\_Latn}, while the remaining bars represent the released German comparison datasets.
    Each bar sums the full GPT-OSS token count of documents satisfying the displayed outcome criterion~\citep{openai2025gptoss}.
    Bars are ordered to pair RefinedGAW-HTML-DE with FineWeb2-DE and RefinedGAW-PDF-DE with FinePDFs-DE; German Commons is a mixed-format contextual reference.
    The leading panel reports total corpus size.
    Every panel uses its own vertical scale, so bar heights are comparable within a panel but not across panels.
    Document-level outcomes can overlap, so yields must not be summed across panels.
    HPLT 3.0-DE is omitted because we did not tokenize it.
    }
    \label{fig:propella-token-yields}
\end{figure*}

The document-weighted comparison in Figure~\ref{fig:propella-comparison} describes prevalence, whereas Figure~\ref{fig:propella-token-yields} reports the absolute amount of text available under each annotation criterion.
For each outcome, we sum the full GPT-OSS token count of every document assigned one of the included labels.
Because propella annotations are document-level and multi-label, these values describe overlapping, selectable token pools rather than disjoint corpus partitions or passage-level measurements.

The full profile shows that absolute yield depends jointly on corpus scale and content concentration.
Although RefinedGAW-HTML-DE has approximately one-sixth as many tokens as FineWeb2-DE, it contributes 53.6 versus 22.3 billion academic-research tokens, 29.7 versus 13.2 billion scientific-technical tokens, and 49.8 versus 46.3 billion advanced-audience tokens.
It remains below FineWeb2-DE for several other outcomes, including analytical, instructional, educational, and reasoning-oriented content.
RefinedGAW-PDF-DE is also smaller than FinePDFs-DE and has lower yields for most outcomes, but contributes 8.7 versus 5.0 billion math-heavy tokens.
The comparison therefore supports property- and format-specific claims: high concentration can offset smaller corpus size for selected content types, but does not imply a uniform ordering across the profile.

\FloatBarrier

\section{Additional Corpus Characterization}
\label{app:additional-characterization}

\subsection{Source-Host Concentration}
We count the number of times a host appears in canonical URLs to examine which hosts contribute how much to RefinedGAW.
A host is a network hostname rather than an institution, so institution-level analyses would require consolidating related subdomains.
In the HTML subset, a few hosts contribute a large share of tokens relative to their number of documents.
The eight highest-yield hosts contribute 33.4\% of tokens but only 4.5\% of documents.
The top host alone contributes 9.3\% of tokens from 0.09\% of documents.
The eight top-hosts in the PDF subset contribute 12.6\% of tokens and 12.1\% of documents.
Half of all tokens come from only 37 HTML hosts, but 162 different PDF hosts.
Therefore, RefinedGAW is not a host-balanced sample, but users can use the retained provenance to cap, reweight, or exclude dominant hosts downstream.

\begin{figure*}[t]
    \centering
    \includegraphics[width=\textwidth]{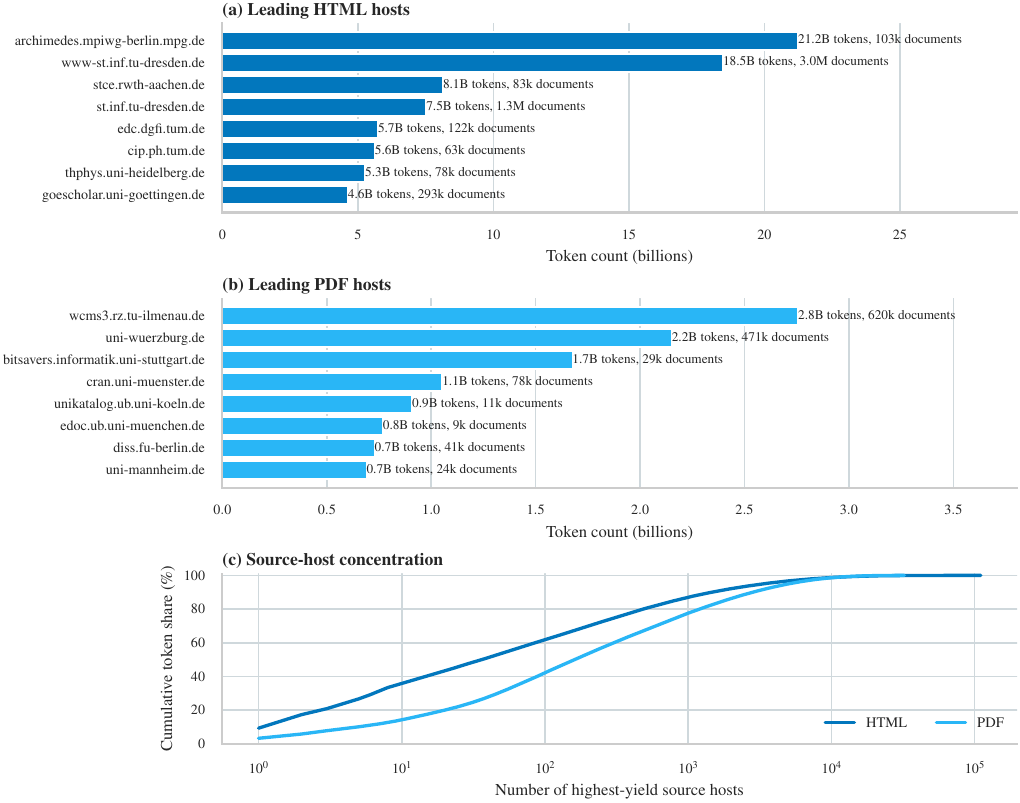}
    \caption{Leading source hosts and source-host concentration in RefinedGAW.
    Panels (a) and (b) report token and document counts for the eight highest-yield hosts in each format.
    Panel (c) accumulates hosts in decreasing order of token yield.
    Each exact-text-deduplicated document is counted once under the host of its canonical retained URL, so the figure describes the indexed collection rather than raw crawl captures.
    Subdomains are treated as separate hosts.}
    \label{fig:source-host-concentration}
\end{figure*}
\FloatBarrier

\subsection{Longitudinal Contribution}

\begin{figure*}[t]
    \centering
    \includegraphics[width=\textwidth]{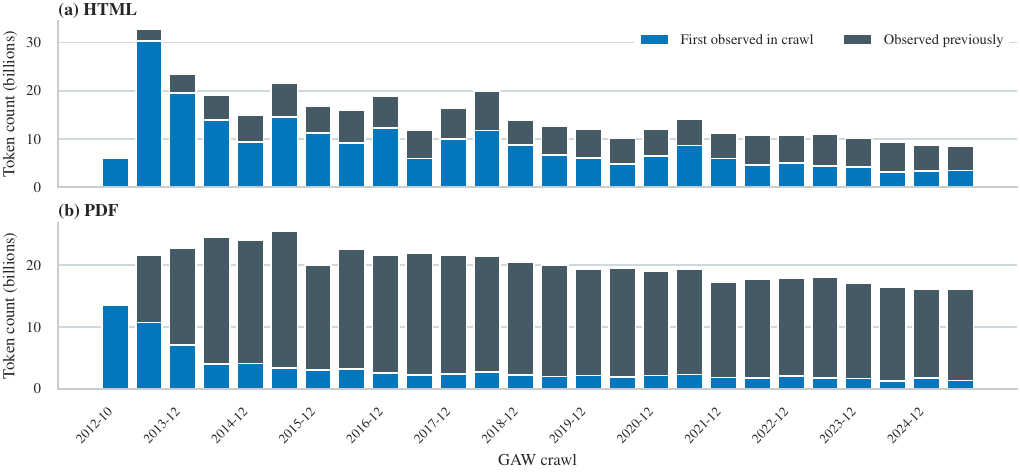}
    \caption{HTML and PDF token yield in each GAW crawl, partitioned by whether an exact-text-deduplicated document first appears in that crawl or was observed previously.
    A recurring document contributes its tokens to every crawl in which it occurs, so the bars describe crawl coverage and are not disjoint partitions of the processed collection.
    First observed denotes the first appearance in GAW and does not establish when the source was published online.
    Tokens are counted with the GPT-OSS tokenizer~\citep{openai2025gptoss}.}
    \label{fig:temporal-contribution}
\end{figure*}

Exact-text deduplication produces one indexed record per unique text, but its provenance still records every crawl in which that text appeared.
Figure~\ref{fig:temporal-contribution} uses this history to separate text first observed in each crawl from exact text already observed in an earlier crawl.
For HTML, the first-observed share falls from 92.4\% of tokens in February 2013 to 40.5\% in June 2025, while the corresponding PDF share falls from 49.4\% to 8.3\%.
Nevertheless, the latest crawl still contributes 3.4 billion previously unseen HTML tokens and 1.3 billion previously unseen PDF tokens.
The provenance therefore preserves both continued additions and a substantial recurring record, especially for PDF.
``First observed'' refers to the archive rather than the publication date of the source, and changes across bars can reflect crawl coverage as well as changes on the web.

\FloatBarrier

\subsection{PDF Length and Extraction Diagnostics}

\begin{figure*}[t]
    \centering
    \includegraphics[width=\textwidth]{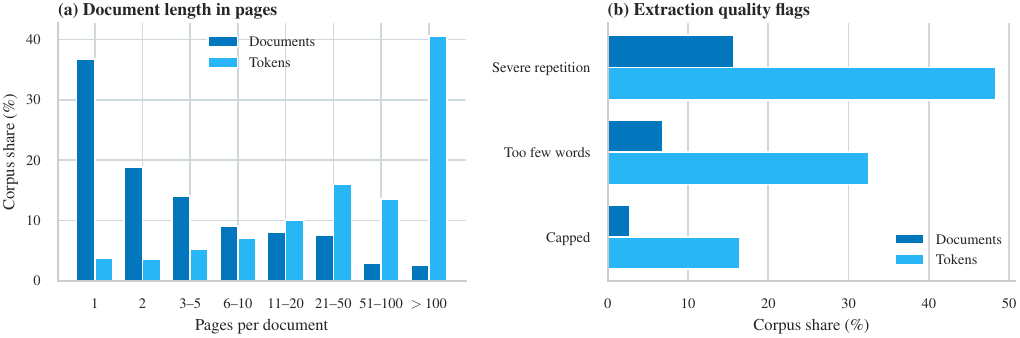}
    \caption{PDF page counts and extraction quality flags.
    Panel (a) contrasts document-weighted and GPT-OSS-token-weighted page-count distributions.
    Panel (b) gives the document and token shares of retained PDFs carrying each extraction flag.
    Quality flags are non-exclusive, and documents without a flag are omitted from panel (b).}
    \label{fig:pdf-extraction-diagnostics}
\end{figure*}

PDF length is highly skewed.
Only 3\% of PDF documents in the final release exceed 100 pages, yet they account for 41\% of PDF tokens.
These documents therefore have a disproportionate effect on corpus composition and token-weighted error analysis.
The non-exclusive extraction flags in Figure~\ref{fig:pdf-extraction-diagnostics}b diverge sharply between the two weightings.
Severe repetition is the most frequent warning, affecting 16\% of documents but 48\% of PDF tokens, and documents carrying no flag make up 80\% of documents yet only 39\% of token mass.
The too-few-words and extraction-capped flags show the same imbalance, affecting 7\% and 3\% of documents but 32\% and 17\% of tokens, respectively.
These flags are pipeline diagnostics rather than human-validated error labels, and multiple flags can apply to one document.
Flagged documents are retained in the release, so extraction-quality selection should consider token-weighted as well as document-weighted prevalence.

\FloatBarrier

\section{Overlap Validation and Reference Intersections}
\label{app:overlap-validation}

\subsection{Candidate Validation and Length Sensitivity}

The textual comparison lowercases text, normalizes numbers and whitespace, removes punctuation and Unicode combining marks, and splits the result on whitespace.
Normalized exact matching hashes the complete normalized document, whereas MinHash uses distinct consecutive five-word shingles from at most the first one million normalized characters.
The DataTrove configuration uses 112 permutations with seed 1, arranged into 14 bands of eight values, and returns a candidate when at least one band matches~\citep{penedo2024datatrove}.
For exact five-shingle Jaccard similarity $s$, the resulting candidate probability is $1-(1-s^8)^{14}$: approximately 5\% at $s=0.5$, 21\% at 0.6, 56\% at 0.7, 92\% at 0.8, and more than 99.9\% at 0.9.
The nominal 0.8 setting therefore favors recall near the target similarity but neither imposes a hard cutoff nor guarantees retrieval.

\begin{figure*}[t]
    \centering
    \includegraphics[width=\textwidth]{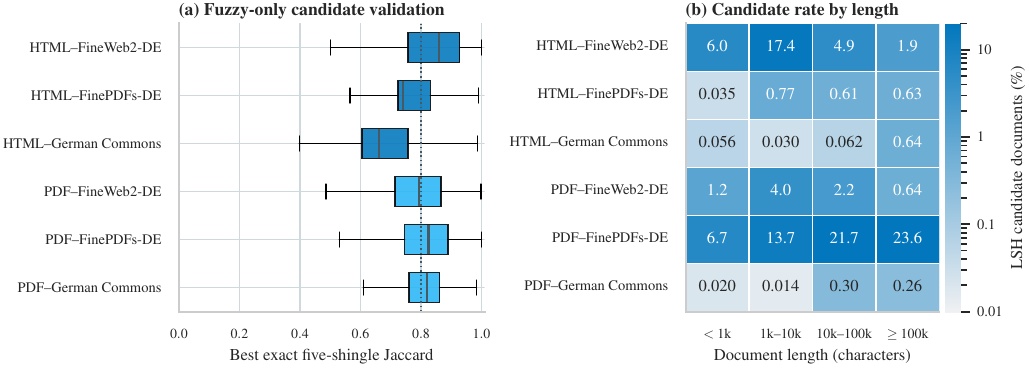}
    \caption{Validation and length sensitivity of MinHash LSH candidates for RefinedGAW-DE query documents.
    Panel (a) gives the exact best five-shingle Jaccard similarity for fuzzy-only candidate samples after restricting query documents to the German subset.
    Boxes show the interquartile range and median, whiskers extend to 1.5 times the interquartile range, and the vertical dotted line marks 0.8 similarity.
    Panel (b) gives corpus-wide candidate rates within RefinedGAW-DE by query-document character length.
    Rows identify the RefinedGAW source format and reference corpus, cell labels give percentages, and color intensity follows a logarithmic scale.}
    \label{fig:overlap-validation}
\end{figure*}

Figure~\ref{fig:overlap-validation}a calibrates the screen on fuzzy-only candidates, which exclude normalized exact matches.
After restricting the validation samples to RefinedGAW-DE, the six directional comparisons contain between 5,331 and 9,949 fuzzy-only pairs each, totaling 53,490.
Only 19\% to 67\% of these candidates per pair reach an exact Jaccard similarity of 0.8.
For the format-matched comparisons, 67\% of RefinedGAW-HTML/FineWeb2-DE candidates and 59\% of RefinedGAW-PDF/FinePDFs-DE candidates reach 0.8, with two-sided 95\% Wilson intervals narrower than two percentage points.
Table~\ref{tab:textual-overlap} must therefore be read as candidate coverage under this configuration rather than verified near-duplicate prevalence.

Candidate coverage also varies with query-document length.
Within RefinedGAW-DE, the HTML/FineWeb2-DE rate peaks at 17\% between 1,000 and 10,000 characters, whereas the PDF/FinePDFs-DE rate increases from 7\% below 1,000 characters to 24\% at 100,000 characters or more.
These opposing patterns show why a single aggregate rate does not describe overlap uniformly across document lengths.

\FloatBarrier

\subsection{Reference-Corpus Intersections}

\begin{figure*}[t]
    \centering
    \includegraphics[width=\textwidth]{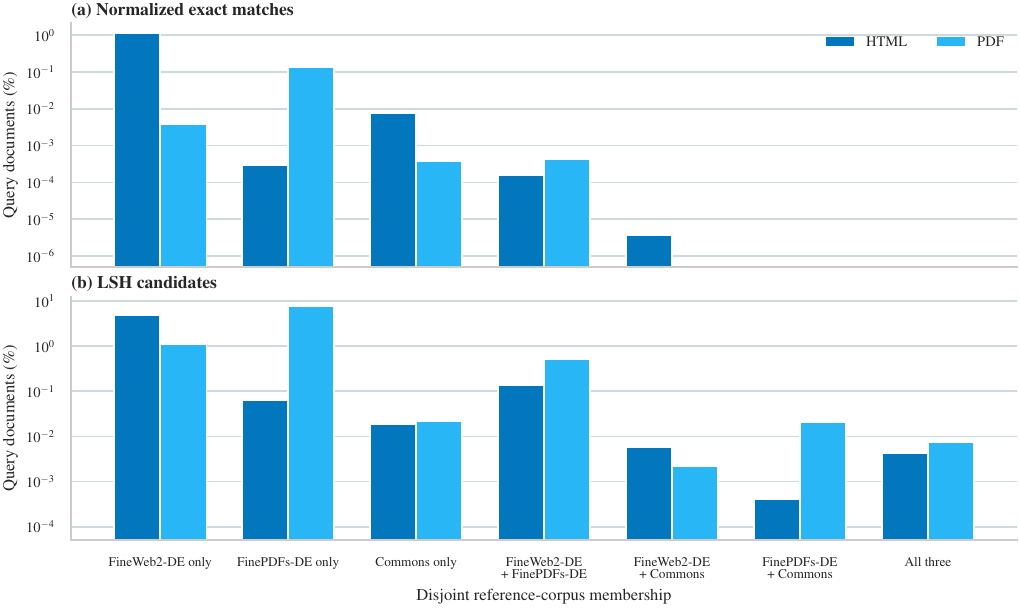}
    \caption{Disjoint intersections of complete-release RefinedGAW exact-match and candidate sets across the three reference corpora.
    Panel (a) gives normalized exact-text matches, while panel (b) gives MinHash LSH candidate sets.
    The vertical axis gives the percentage of documents in the corresponding RefinedGAW source format, using a logarithmic scale.
    Each category contains documents assigned to exactly the displayed set of reference corpora, and an absent bar denotes zero documents.
    The LSH panel describes candidate-set membership and does not verify a fixed similarity threshold.}
    \label{fig:overlap-intersections}
\end{figure*}

A RefinedGAW document can match more than one reference corpus, so the pairwise rates cannot be added.
Figure~\ref{fig:overlap-intersections} resolves this ambiguity by partitioning each union into disjoint memberships.
The normalized exact comparison hashes complete documents after the same lowercasing, number, punctuation, diacritic, and whitespace normalization used for MinHash.
Format-matched-only memberships account for 99.3\% of the HTML normalized-exact union and 96.8\% of the PDF union, corresponding to 1.104\% of all HTML documents matching only FineWeb2-DE and 0.139\% of all PDFs matching only FinePDFs-DE.
The candidate sets overlap more broadly: 151,207 HTML documents and 54,616 PDFs are candidates in both FineWeb2-DE and FinePDFs-DE but not German Commons, representing 2.6\% and 5.6\% of their respective candidate unions.
Memberships involving German Commons remain rare under either comparison.

\FloatBarrier

\section{Rehydration Diagnostics}
\label{app:rehydration-diagnostics}

The equal-sized samples drawn from the Top 10\% and Other 90\% deliberately oversample higher-ranked documents.
Unless a result is reported separately for these groups, we therefore combine their estimates with weights of 0.1 and 0.9 to represent their shares of the format-specific index.
A record is \textit{Downloaded} when at least one URL in its deterministic primary-and-alternative sequence returns an eligible public resource under the tool's access and size constraints.
The \textit{Extracted} stage additionally requires production-pipeline text of more than 300 characters for HTML or nonempty assembled OCR text for PDF; the latter stage does not reapply the release's minimum-length PDF filter.
The similarity criterion requires a Jaccard score of at least 0.8 over distinct consecutive five-word shingles after lowercasing, number and whitespace normalization, and removal of punctuation and diacritics.
Exact match requires the new extraction to equal the archived text character for character, so a changed date, timestamp, or footer is enough to fail, and neural OCR running on GPUs can additionally vary with hardware and batching.
These four outcomes are nested, and each group-specific rate uses all 5,000 sampled records in that group as the denominator.

\begin{figure*}[t]
    \centering
    \includegraphics[width=\textwidth]{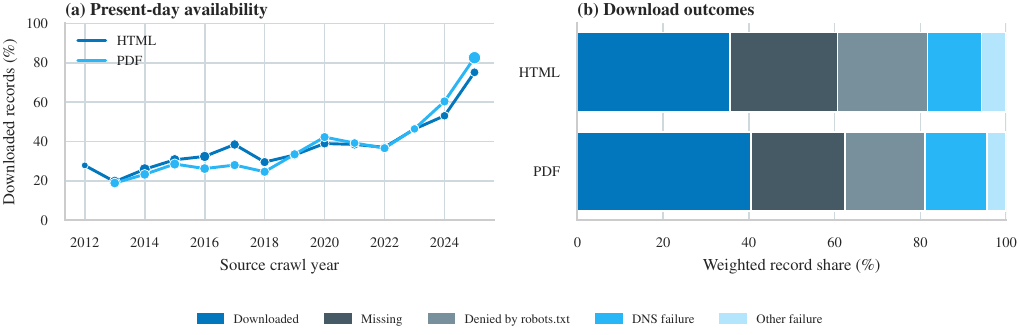}
    \caption{Present-day source availability in the two-group rehydration sample.
    Panel (a) gives download rates by source crawl year after weighting estimates from the Top 10\% and Other 90\% groups by their estimated within-year population shares; marker area is proportional to the estimated share of the corresponding format index represented by that year.
    Panel (b) gives full-index download outcomes after combining the Top 10\% and Other 90\% samples with weights 0.1 and 0.9.
    Other failure combines categories that each account for less than 2\% of the corresponding format index.}
    \label{fig:rehydration-diagnostics}
\end{figure*}

Figure~\ref{fig:rehydration-diagnostics} separates source age from present-day download outcomes.
Weighted availability rises to 75\% for HTML and 83\% for PDF in 2025, but the year-specific estimates remain descriptive because the number of sampled records varies by year.
At the full-index level, missing resources, current \texttt{robots.txt} exclusions, and DNS failures account for 25\%, 21\%, and 13\% of HTML records and 22\%, 19\%, and 14\% of PDF records, respectively.
Together, these three categories explain 91\% of unsuccessful HTML downloads and 93\% of unsuccessful PDF downloads.

Source recency also explains part, but not all, of the association between quality rank and availability.
The unadjusted Top 10\% minus Other 90\% download gap is 10.1 percentage points for HTML and 11.4 points for PDF.
Standardizing both groups to the estimated crawl-year distribution of the corresponding full index reduces these gaps to 6.3 and 7.6 points, respectively.
Among successfully extracted records, median normalized five-shingle similarity is 0.809 for Top 10\% HTML versus 0.278 for Other 90\% HTML, and 0.982 versus 0.939 for PDF.
These conditional similarities describe fidelity among successful extractions and should not be combined directly with the unconditional rates in Table~\ref{tab:url-reachability}.

\FloatBarrier

\section{Quality-Score Construction and Diagnostics}
\label{app:quality-score}

We assign a quality score (integer between 0 and 132) to each document as a convenience ranking rather than a learned or human-calibrated estimate of document quality.
It combines the seven categorical propella properties as detailed in Table~\ref{tab:quality-score}.
The score gives double weight to content quality, educational value, reasoning indicators, and commercial bias.
Our release sorts documents by decreasing score, with the document identifier as a deterministic tie-breaker (ascending).
The quality score does not include language, academic sector, technical content, audience level, content safety, or PII annotations.

\begin{table*}[t]
    \centering
    \small
    \setlength{\tabcolsep}{5pt}
    \begin{tabular}{@{}p{0.17\textwidth}cp{0.72\textwidth}@{}}
        \toprule
        Property & Weight & Categories from lowest to highest component value \\
        \midrule
        Content quality & 2 & \texttt{unacceptable}, \texttt{poor}, \texttt{adequate}, \texttt{good}, \texttt{excellent} \\
        Educational value & 2 & \texttt{none}, \texttt{minimal}, \texttt{basic}, \texttt{moderate}, \texttt{high} \\
        Reasoning indicators & 2 & \texttt{none}, \texttt{minimal}, \texttt{basic\_reasoning}, \texttt{explanatory}, \texttt{analytical} \\
        Commercial bias & 2 & \texttt{pure\_marketing}, \texttt{heavy}, \texttt{moderate}, \texttt{minimal}, \texttt{none} \\
        Information density & 1 & \texttt{empty}, \texttt{thin}, \texttt{moderate}, \texttt{adequate}, \texttt{dense} \\
        Content integrity & 1 & \texttt{severely\_degraded}, \texttt{fragment}, \texttt{mostly\_complete}, \texttt{complete} \\
        Content ratio & 1 & \texttt{minimal\_content}, \texttt{mostly\_navigation}, \texttt{mixed\_content}, \texttt{mostly\_content}, \texttt{complete\_content} \\
        \bottomrule
    \end{tabular}
    \caption{Components of the RefinedGAW quality score.
    Categories are listed in ascending component order; commercial bias is therefore reverse-coded so that less commercial text receives a larger component value.}
    \label{tab:quality-score}
\end{table*}

\FloatBarrier

\end{document}